\documentclass{article}
\usepackage{spconf,amsmath,graphicx}
\usepackage{float}
\newcommand{\comment}[1]{}

\title{Anatomical Data Augmentation For CNN based Pixel-wise Classification}

\name{Avi Ben-Cohen$^{\star}$ \quad Eyal Klang$^{\dagger}$ \quad Michal Marianne Amitai$^{\dagger}$ \quad Jacob Goldberger$^{\ddagger}$ \quad Hayit Greenspan$^{\star}$}

\address{$^{\star}$ Tel Aviv University, Faculty of Engineering, Department of Biomedical Engineering,\\ Tel Aviv 69978, Israel\\
$^{\dagger}$ Sheba Medical Center, Diagnostic Imaging Department, Abdominal Imaging Unit, \\ affiliated to Sackler school of medicine Tel Aviv University, Tel Hashomer 52621, Israel\\
$^{\ddagger}$ Bar-Ilan University, Faculty of Engineering, Ramat-Gan, 5290002, Israel}

\begin{document}
%
\maketitle
\begin{abstract}

          In this work we propose a method for anatomical data augmentation that is based on using slices of computed tomography (CT) examinations that are adjacent to labeled slices  as another resource of labeled data for training the network.
   The extended labeled data is used to train a  U-net network for a pixel-wise classification into
      different hepatic lesions and normal liver tissues.
               Our dataset contains CT examinations from 140 patients with 333 CT images annotated by an expert radiologist.  We tested our approach and compared it to the conventional training process. Results indicate superiority of our method. Using the anatomical data augmentation we achieved an improvement of $3\%$ in the success rate, $5\%$ in the classification accuracy, and $4\%$ in Dice.
\end{abstract}
\section{Introduction}
\label{sec:intro}
Convolutional Neural Networks (CNN) have shown outstanding performance in visual object recognition and image classification in general imagery as well as more recently in the medical domain \cite{Greenspan}. Deep learning is a data driven approach and although in recent years a few datasets in medical imaging have become public, for most tasks there is still a lack of annotated data.
The conventional approach to deal with this problem is to augment the existing data. Most data augmentation approaches include simple modifications of the images such as scale, translation, and rotation when relevant to the task. Applying these  modifications can be less effective when dealing with medical data where there is a lack of labeled data and the anatomical changes are much more diverse.

In the current work we focus on liver lesion analysis.  Computed tomography (CT) is commonly used for detection and classification, as well as follow-up (FU) of liver lesions \cite{hopper2000body}. Current radiological practice is to manually analyze the liver. The liver can include multiple lesions from several types which makes it a complex task. In order to facilitate the task, the images are acquired before and after intravenous injection of a contrast agent. In most cases, the optimal detection of liver lesions is in the portal phase, which is 60-80 seconds post injection.  

Several steps  in automating the liver analysis can be found in recent literature: including liver segmentation, lesion detection, lesion categorization and follow-up. Among these,  the lesion {\it{segmentation}} task  has attracted the greatest deal of attention in recent years, being the focus of a couple of challenges (ISBI 2017 and MICCAI 2017 LITS challenges \cite{LiTS}). The objective of these challenges is to find a segmentation mask similar to the one manually circumscribed by experts. In the challenge dataset, most lesions are malignant lesions. In practice, it is also important to separate between malignant and benign lesions.

In this paper we focus on pixel-wise classification of liver CT images that include both benign and malignant hepatic lesions.
Our pixel-wise classification model accepts a liver CT image and outputs per-pixel classification into 6 classes: background, interior liver, liver boundary, metastasis, hemangioma, and cyst\footnote{The liver boundary class was added to help the network model distinguish between lesions and liver boundary tissue as suggested in \cite{frid}.}.
A main challenge in pixel-wise classification is the lack of sufficient training samples. In this work we suggest a new procedure to augment existing training data, which we term anatomical-based augmentation:   

We first train a classifier using  an existing set of labeled slices. We next use the network to classify unlabeled  slices that are adjacent to the labeled slices. Finally, we retrain the network using these slices as another resource of labeled data. Our motivation is based on the fact that slices that are adjacent to labeled slices are expected to have similar anatomical structure and therefore, given the labeled slices,  we can accurately classify them. On the other hand, every slice is different and therefore using these slices increases the diversity of the training data.  We show that this anatomical based augmentation strategy  can improve the network's performance in terms of classification and segmentation of medical data.

\section{Anatomical data augmentation}
\label{sec:anatomical}

Deep learning based methods commonly utilize supervised learning. The challenge with supervised learning is the need for a large amount of labeled data. Collecting this data can be very expensive especially for medical applications. 
In semi-supervised learning techniques, labeled data is used as well as unlabeled data in the learning scheme.
Most previous deep learning approaches have used unlabeled data only for pre-training which is
a way to initialize the weights of the network followed by normal supervised learning. The standard pre-training strategy is based on a greedy layer-wise procedure using either Restricted Boltzmann Machines (RBM) \cite{Hinton2006} or noisy autoencoders \cite{Bengio2007}.
 Ladder networks \cite{rasmus} is a way to combine supervised learning with unsupervised learning in deep neural networks by explicitly incorporating the unlabeled data into the cost function
that is optimized in the training step.
In this study we propose a semi-supervised learning scheme that uses unlabeled data as another way of augmentation.

Our dataset included segmentation masks of the liver and liver lesions as well as their type (metastasis, hemangioma, or cyst). Only a few slices in each CT scan were labeled so the dataset was overall relatively small. Hence, we wanted to make use of some of the unlabeled dataset to improve the network's performance. Since adjacent CT slices are relatively similar in their characteristics, we took two adjacent slices above and below each training image and used them for training. Instead of using the same label map as the center slice, we generated a new label map for each adjacent slice using a trained model. We follow a similar concept as suggested in \cite{lee} which showed the benefit of using unlabeled data in deep neural networks. Our proposed framework includes the following steps:
\begin{enumerate}
\item Train the network model using the labeled training set.
\item Run the trained model on the unlabeled adjacent slices and save the per-pixel classification results.
\item Re-train the network model using the labeled data as before and add the unlabeled data modifying its generated label map to include the uncertainty.
\end{enumerate}
Note that we are not simply propagating the labels of the annotated slice to the adjacent slice.   We show in section \ref{sec:results} that this n\"aive approach yields inferior results. This is due to the fact that although the slices are similar, they are not the same.
In our solution we used a weighted cross entropy loss function $L$ (equation \ref{eq:loss}), to balance the learning process, with weights $\omega^{c}$ that are inversely proportional to the ratio of pixels in each class.
\begin{equation}
\label{eq:loss}
    L =-  \sum\limits_{i=1}^N \omega_i^{c}  \left[ \hat{P_i^{c}} \log P_i^{c} \right]
\end{equation}
Where $P_i^{c}$ denotes the probability of a pixel $i$ belonging to each class $c$ and $\hat{P_i^{c}}$ represents the ground truth. When using the labeled training set, $\hat{P_i^{c}}$ was 1 for the correct class and 0 for other classes. Using the unlabeled dataset  $\hat{P_i^{c}}$ we propose to use a range of values,  $\gamma \in (0.5,1)$ instead of 1. This reflects the uncertainty in this dataset compared to the labeled one. The value of $\gamma$ was chosen manually to be $0.7$ with no optimization process although different values should be tested.

\section{Experiment setup}
\label{sec:setup}
\subsection{Dataset}
\begin{figure}
\centering
\includegraphics[height=9 cm]{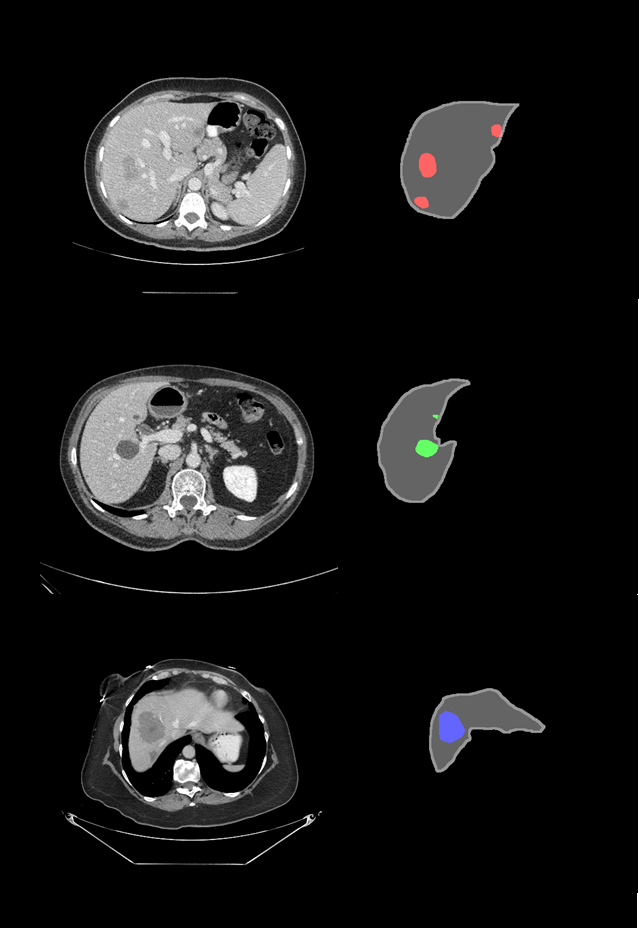}
\caption{CT input examples (left); Labeled lesions and liver (right). In red - metastasis; green - cyst; blue - hemangioma.}
\label{fig:dataset}
\end{figure}

The data used in the current work included CT scans from the Sheba Medical Center. Different CT scanners were used with 0.71-1.17 mm pixel spacing and 1.25-5 mm slice thickness. Markings of the entire liver and hepatic lesions boundary were conducted by an expert radiologist.
The dataset included 333 CT slices with annotations taken from 140 patients. It was divided into a training set with 225 images and testing set with 108 images. Table \ref{tab:dataset} shows the number of images in each class (metastasis, hemangioma, cyst, and healthy) that were used in the training and testing dataset. Fig. \ref{fig:dataset} shows an example of the data used in this work.

\begin{table}
\centering
\caption{The number of images divided into classes used for the training and testing datasets.}
\label{tab:dataset}
\begin{tabular}{|l|cccc|}
\hline
         & Metastasis & Hemangioma & Cyst & Healthy \\ \hline
Training & 64         & 48         & 51   & 62      \\ \hline
Testing  & 30         & 23         & 24   & 31      \\ \hline
Total    & 94         & 71         & 75   & 93      \\ \hline
\end{tabular}
\end{table}

\subsection{Network architecture}
Several previous works have shown the superiority of fully convolutional networks for liver lesion detection and segmentation \cite{Ben-Cohen,Christ,Dou}. Hence, we chose our network model to be “U-Net” based \cite{Ronneberger}. It includes an encoder and a decoder, where the final output of the network is in the same size of the image.

Let $C_{k,s}$ denote a Convolution-ReLU layer and $CB_{k,s}$ denote a Convolution-ReLU-BatchNorm with $k$ filters of size $s\times s$, $CD_{k,s}$ denotes a Convolution-ReLU-Dropout layer and $CBD_{k,s}$ denotes a Convolution-ReLU-BatchNorm-Dropout with a dropout rate of 50\%.  $CBP_{k,s}$ denotes a Convolution-ReLU-BatchNorm-Maxpool with pooling size of $2\times 2$, and $U_f$ denotes an Upsampling (deconvolution) layer by factor $f$. The``U-Net" encoder:\\
$C_{64,3} \rightarrow CBP_{64,3} \rightarrow C_{128,3} \rightarrow CBP_{128,3} \rightarrow C_{256,3} \rightarrow CBP_{256,3} \rightarrow C_{512,3} \rightarrow CBP_{512,3} \rightarrow C_{1024,3} \rightarrow CB_{1024,3}$.
The ``U-Net" decoder:\\
$U_{2} \rightarrow C_{512,3} \rightarrow CB_{512,3} \rightarrow U_{2} \rightarrow C_{256,3} \rightarrow CBD_{256,3} \rightarrow U_{2} \rightarrow C_{128,3} \rightarrow CBD_{128,3} \rightarrow U_{2} \rightarrow CD_{64,3} \rightarrow C_{64,3} \rightarrow C_{6,1}$.\\
The ``U-net" includes skip connections between every second layer $i$ in the encoder and layer $n-i-1$ in the decoder, where $n$ is the total number of layers. The skip connections concatenate activations from layer $i$ to layer $n-i-1$.

\subsection{Online data augmentation}
\label{sec:augmentation}
To assist the network model to be more robust to  liver lesions variability in terms of location and scaling,  online data augmentation was performed with uniform sampling of scale [-0.9,1.1] and translations [-25,25].  Two augmentations (with different scales and translations) were performed in each epoch for each training set image during the training process.

\subsection{Post processing}
Lesion candidates with an area smaller than $1 cm^2$ were ignored. In addition, we trained a separate model for only liver segmentation to refine the unlabeled dataset label maps.

\section{Experiments and Results}
\label{sec:results}
The objective  was to test whether  the additional training with the unlabeled dataset improved the segmentation and classification performance as  compared to the original model.
We used the following measurements to evaluate each model's performance:
\textbf{Success}: number of images in which the lesion's ground truth segmentation overlaps the model's segmentation divided by the number of images in the test set.
 \textbf{Dice1}: the average Dice segmentation measurement between lesions and not lesions, calculated per image where there is an overlap.
\textbf{Dice2}: the average Dice segmentation measurement between lesions and not lesions, calculated per image including cases with no overlap.
\textbf{ACC}: each image was classified based on the majority class (between the different lesion classes) and the accuracy in classification was measured.

In the first experiment, we explore  the benefit of training with the additional unlabeled data as compared to conventional data augmentation. We term the model that was trained using labeled data as the {\it baseline model}.  We define an {\it extended baseline model} in which we increase the amount of conventional data augmentation, to match the exact number of training samples as in the unlabeled data case. 
In the presented results we use the following: For the baseline model  we used  2 conventional data augmentations per image. Similarly, 2 conventional data augmentations are conducted for each additional adjacent slice in the anatomical data augmentation model. To get a similar number of augmented images, we use 6 conventional data augmentations per image  in the extended baseline model.  

Table \ref{tab:results} shows the results, comparing the baseline model, the extended baseline model 
and our proposed model, in terms of Dice, Success rate, and classification accuracy. Our proposed method obtained an improvement of $4\%$ in Dice1, $5\%$ in Dice2, $3\%$ in the success rate, and $5\%$ in the classification accuracy compared to the baseline model training. When using an extended amount of data augmentation, the Dice measure and the classification accuracy improved but were still lower than our proposed method's performance.

\begin{figure}[t!]
\centering
\includegraphics[height=9 cm]{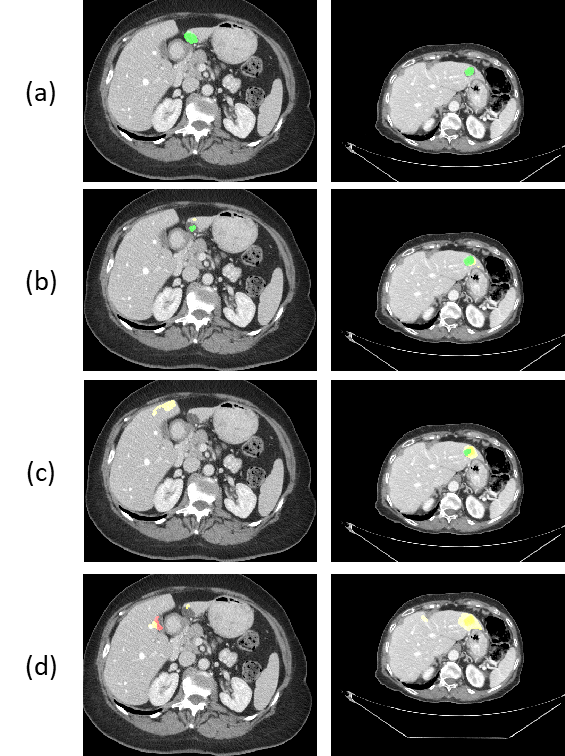}
\caption{Qualitative results: a) Ground truth; b) Model trained using our proposed method; c) Using extended data augmentation; d) Baseline model.  red - metastasis, green - cysts, and yellow - hemangioma.}
\label{fig:results}
\end{figure}

\begin{table}
\centering
\caption{Evaluation measures for segmentation and classification comparing the baseline model trained with labeled data, trained with extended data augmentation, and trained using our proposed method with anatomical data augmentation (in bold).}
\medskip
\label{tab:results}
\begin{tabular}{|c|cccc|}
\hline
         & Dice1 & Dice2 & Success & ACC \\ \hline
Anatomical & \textbf{83}         & \textbf{66}         & \textbf{80}   & \textbf{61}      \\ \hline
Extended  & 82         & 63         & 77   & 57      \\ \hline
Baseline    & 79         & 61         & 77   & 56      \\ \hline
\end{tabular}
\end{table}

\begin{table}
\centering
\caption{Evaluation measures for segmentation and classification comparing the baseline model trained with anatomical data augmentation with $\gamma=0.7$ (in bold), $\gamma=1.0$, and with the label map taken from the neighbor slice that had ground truth annotations (no label map generation).}
\medskip
\label{tab:comparison}
\begin{tabular}{|c|cccc|}
\hline
         & Dice1 & Dice2 & Success & ACC \\ \hline
$\gamma=0.7$ & \textbf{83}         & \textbf{66}         & \textbf{80}   & \textbf{61}      \\ \hline
$\gamma=1.0$  & 79         & 62         & 79   & 58      \\ \hline
Neighbor labels    & 86         & 64         & 75   & 59      \\ \hline
\end{tabular}
\end{table}

Fig. \ref{fig:results} shows  qualitative results for two input cases, when using the different training approaches. In both cases, adding the anatomical data augmentation in the training process (b) provided better results in terms of segmentation and classification.

Table \ref{tab:comparison} shows the comparison of our proposed anatomical data augmentation with several variants of it.  First, 
we tested the model using $\gamma=1.0$ as in regular training process with labeled data. 
The second variation is in the labeling of the neighboring slice maps. Instead of generating them (from the baseline model), we take  the same label map as in the ground truth label map of the labeled slice ($\gamma=0.7$). 
The results show  that our approach provided better results in most terms except Dice1 which achieved a higher score using the neighbor label map but with lower Dice2, success, and ACC.

\section{Conclusion}

  In this study we showed a novel approach for using unlabeled data in medical images to improve CNN based pixel-wise classification. We believe that the success of using  unlabeled adjacent slices for augmentation is due to the anatomical relevancy, thus in effect utilizing the medical context to our benefit.
  Our approach is simple and easy to implement. In the future, we plan to conduct  experiments to test the robustness of the scheme to additional more-complex data augmentations, check the sensitivity to number of slices added, and to explore the generalizability to  different datasets and tasks. \\

\textbf{Acknowledgement} This research was supported by the Israel Science Foundation (grant No. 1918/16).

\small
\bibliographystyle{IEEEbib}
\bibliography{refs}

\begin{thebibliography}{10}

\bibitem{Greenspan}
H.~Greenspan, B.~van Ginneken, and R.M. Summers,
\newblock ``Guest editorial deep learning in medical imaging: Overview and
  future promise of an exciting new technique,''
\newblock {\em IEEE Transactions on Medical Imaging}, vol. 35, no. 5, pp.
  1153--1159, 2016.

\bibitem{hopper2000body}
K.~D Hopper, K.~Singapuri, and A.~Finkel,
\newblock ``Body {CT} and oncologic imaging 1,''
\newblock {\em Radiology}, vol. 215, no. 1, pp. 27--40, 2000.

\bibitem{LiTS}
P.F. Christ,
\newblock {\em LiTS- Liver Tumor Segmentation Challenge, ISBI17 and MICCAI17},
  2017.

\bibitem{frid}
M.~Frid-Adar, I.~Diamant, E.~Klang, M.~Amitai, J.~Goldberger, and H.~Greenspan,
\newblock ``Modeling the intra-class variability for liver lesion detection
  using a multi-class patch-based cnn,''
\newblock in {\em International Workshop on Patch-based Techniques in Medical
  Imaging}. Springer, 2017, pp. 129--137.

\bibitem{Hinton2006}
G.E. Hinton, S.~Osindero, and Y.W. Teh,
\newblock ``A fast learning algorithm for deep belief nets,''
\newblock {\em Neural computation}, vol. 18, no. 7, pp. 1527--1554, 2006.

\bibitem{Bengio2007}
Y.~Bengio, P.~Lamblin, D.~Popovici, and H.~Larochelle,
\newblock ``Greedy layer-wise training of deep networks,''
\newblock in {\em Advances in neural information processing systems}, 2007, pp.
  153--160.

\bibitem{rasmus}
A.~Rasmus, M.~Berglund, M.~Honkala, H.~Valpola, and T.~Raiko,
\newblock ``Semi-supervised learning with ladder networks,''
\newblock in {\em Advances in Neural Information Processing Systems}, 2015, pp.
  3546--3554.

\bibitem{lee}
D.H. Lee,
\newblock ``Pseudo-label: The simple and efficient semi-supervised learning
  method for deep neural networks,''
\newblock in {\em Workshop on Challenges in Representation Learning, ICML},
  2013, vol.~3, p.~2.

\bibitem{Ben-Cohen}
A.~Ben-Cohen, I.~Diamant, E.~Klang, M.~Amitai, and H.~Greenspan,
\newblock ``Fully convolutional network for liver segmentation and lesions
  detection,''
\newblock in {\em International Workshop on Large-Scale Annotation of
  Biomedical Data and Expert Label Synthesis}. Springer, 2016, pp. 77--85.

\bibitem{Christ}
P.F. Christ, M.A.E. Elshaer, F.~Ettlinger, S.~Tatavarty, M.~Bickel, P.~Bilic,
  M.~Rempfler, M.~Armbruster, F.~Hofmann, M.~D'Anastasi, et~al.,
\newblock ``Automatic liver and lesion segmentation in {CT} using cascaded
  fully convolutional neural networks and 3d conditional random fields,''
\newblock in {\em International Conference on Medical Image Computing and
  Computer-Assisted Intervention}. Springer, 2016, pp. 415--423.

\bibitem{Dou}
Q.~Dou, H.~Chen, Y.~Jin, L.~Yu, J.~Qin, and P.A. Heng,
\newblock ``3d deeply supervised network for automatic liver segmentation from
  {CT} volumes,''
\newblock in {\em International Conference on Medical Image Computing and
  Computer-Assisted Intervention}. Springer, 2016, pp. 149--157.

\bibitem{Ronneberger}
O.~Ronneberger, P.~Fischer, and T.~Brox,
\newblock ``U-net: Convolutional networks for biomedical image segmentation,''
\newblock in {\em International Conference on Medical Image Computing and
  Computer-Assisted Intervention}. Springer, 2015, pp. 234--241.

\end{thebibliography}

\end{document}